\documentclass[11pt]{article}

\usepackage[final]{acl}

\usepackage{times}
\usepackage{algorithm}
\usepackage{algorithmic}
\usepackage{latexsym}
\usepackage{tabularx}
\usepackage{colortbl}
\usepackage{microtype}
\usepackage{graphicx}
\usepackage{subfigure}
\usepackage{booktabs} 
\usepackage{bm}
\usepackage{hyperref}

\usepackage{algorithm}
\usepackage{algorithmic}
\usepackage{amsmath}
\usepackage{booktabs}
\usepackage{multirow} 
\usepackage{makecell}

\usepackage{newfloat}
\usepackage{listings}
\DeclareCaptionStyle{ruled}{labelfont=normalfont,labelsep=colon,strut=off} 
\floatstyle{ruled}
\newfloat{listing}{tb}{lst}{}
\floatname{listing}{Listing}
\usepackage{tcolorbox}

\newtcolorbox{myprompt_single}[2][]
{
    colframe=black,      
    colback=gray!20,     
    boxrule=1pt,         
    arc=4pt,             
    left=10pt,           
    right=10pt,
    top=10pt,            
    bottom=10pt, 
    title=#2
}

\newtcolorbox{myprompt_double}[2][]
{
    colframe=black,      
    colback=gray!20,     
    boxrule=1pt,         
    arc=4pt,             
    left=10pt,           
    right=10pt,
    top=10pt,            
    bottom=10pt, 
    width=\textwidth,
    title=#2
}

\usepackage{amsmath}
\usepackage{amssymb}
\usepackage{mathtools}
\usepackage{amsthm}

\usepackage[capitalize,noabbrev]{cleveref}

\theoremstyle{plain}
\newtheorem{theorem}{Theorem}[section]
\newtheorem{note}[theorem]{Note}

\theoremstyle{definition}

\theoremstyle{note}

\usepackage[textsize=tiny]{todonotes}
\usepackage[T1]{fontenc}

\usepackage[utf8]{inputenc}

\usepackage{microtype}

\usepackage{inconsolata}

\usepackage{graphicx}
\usepackage{wrapfig}
\title{VizRAG: Enhancing Retrieval-Augmented Generation with Hypergraph Visualization}

\author{
\textbf{Yanbin Wei}$^{1,2}$\thanks{Equal contribution.} \quad
\textbf{Yang Chen}$^{1}$\footnotemark[1] \quad
\textbf{Renling Gan}\footnotemark[1] \quad
\textbf{Ziru Liu}$^{3}$ \quad
\textbf{Xinyu Fu}$^{3}$ \quad \\
\textbf{Chun Kang}$^{4}$ \quad
\textbf{Ning Lu}$^{1,2}$ \quad 
\textbf{Rui Liu}$^{3}$\footnotemark[2] \quad
\textbf{Yu Zhang}$^{1}$\footnotemark[2] \quad
\textbf{James Kwok}$^{2}$ \\
$^1$Southern University of Science and Technology\quad\\
$^2$Hong Kong University of Science and Technology\\
$^3$Huawei Research \quad
$^4$Beihang University\\
}

\begin{document}
\maketitle
\begin{abstract}
Hypergraph-based Retrieval-Augmented Generation (RAG) systems surpass traditional graph-based approaches by organizing complex $n$-ary atomic facts among entities, rather than relying solely on binary relationships. Despite the advancements in multimodal large language models (MLLMs) with enhanced visual capabilities, current hypergraph-based RAG frameworks predominantly restrict knowledge retrieval and reconstruction to a unimodal, text-centric paradigm. This limitation prevents them from fully leveraging the powerful visual perception capabilities of modern MLLMs. To address this gap, we systematically explore the integration of hypergraph awareness in RAG systems through visual cues. By incorporating visual representations of hypergraphs into the RAG pipeline, we introduce VizRAG, the first RAG system to support visual hypergraph structure awareness. Experimental results demonstrate that VizRAG significantly outperforms strong baselines, validating the promising potential of hypergraph visualization as a novel approach for RAG systems.
\end{abstract}

\section{Introduction}
Retrieval-Augmented Generation (RAG) has become a key framework for improving LLM performance in knowledge-intensive tasks like question answering, document analysis, and intelligent assistance \cite{lewis2020retrieval, gao2023retrieval, li2024structrag}. By leveraging external knowledge sources, RAG reduces hallucinations in LLMs, enhancing both the reliability and quality of generated outputs. Moreover, RAG enables seamless integration of private or domain-specific knowledge bases, allowing LLMs to adapt effectively to specialized contexts.

However, conventional RAG systems predominantly utilize chunk-level retrieval based on vector similarity \cite{asai2023self, yang2024crag}. While this approach achieves reasonable effectiveness, it often overlooks inter-chunk relationships and hierarchical semantic structures. As a result, retrieved content may lack coherence and exhibit weak connectivity, limiting the model's ability to structurally interpret knowledge, a critical factor for producing context-rich and coherent responses.

To address these shortcomings, recent studies have explored graph-based structures within RAG frameworks. For instance, GraphRAG \cite{edge2024local} and LightRAG \cite{guo2024lightrag} utilize knowledge graphs to enhance entity-level indexing and retrieval by explicitly modeling semantic relationships. Hyper-RAG \cite{feng2025hyper} and Cog-RAG \cite{hu2026cog} further extend this paradigm by using hypergraphs to capture higher-order relationships among multiple entities, offering greater expressiveness than traditional graphs. However, the approaches remain predominantly text-centric, restricting knowledge retrieval and reasoning to unimodal paradigms and failing to leverage advanced visual perception capabilities of modern multimodal large language models (MLLMs).

In contrast, humans often rely on visual intuition to perceive and reason about complex structures, such as loops, clusters, or sets, within graphical representations. This cognitive advantage suggests that integrating visualized hypergraphs into the RAG process could complement text-based retrieval with multimodal benefits, such as intuitive reasoning shortcuts and enriched structural understanding. Inspired by this insight, we systematically analyze the \textbf{advantages (Sec.~\ref{sec:adv})} and \textbf{feasibility (Sec.~\ref{sec:fea})} of using hypergraph visualization to enhance RAGs, to reveal the untapped potential of hypergraph visualization as a \textbf{novel direction}.

Nevertheless, incorporating hypergraph visualization into RAG systems poses unique challenges (Sec.~\ref{sec:cha}), including \textbf{visual congestion} and \textbf{rendering bias}. Addressing these challenges is essential to fully realize the potential of visual hypergraphs in RAG systems. To this end, we propose \textbf{HyperViz} (Sec.~\ref{sec:hyperviz}), a toolkit that provides scientifically grounded solutions to these challenges, making hypergraph visualization an effective and accessible plugin for RAG systems. 

Building on \textbf{HyperViz},  we introduce \textbf{VizRAG} (Sec.~\ref{sec:vizrag}), a novel RAG framework that integrates hypergraph visualization to enhance the retrieval and reasoning capabilities of MLLMs. By combining visual hypergraph awareness deeply into the retrieval strategies, \textbf{VizRAG} distinguishes itself from the prior works and significantly improves the coherence and quality of generated responses over strong baselines in extensive experiments.

Our contributions are summarized as follows:
\begin{itemize}
    \item We introduce \textbf{hypergraph visualization} as a novel and promising direction for enhancing RAG systems, systematically analyzing its potential, feasibility, and challenges.
    \item We develop \textbf{HyperViz}, a plug-and-play, well-encapsulated toolkit that addresses key challenges in hypergraph visualization.
    \item We propose \textbf{VizRAG},  a visualization-driven RAG framework that incorporates hypergraph visual awareness into the entire RAG pipeline, enabling multimodal retrieval and reasoning.
    \item Extensive experiments show that \textbf{VizRAG} significantly outperforms strong RAG baselines on multiple knowledge-intensive tasks, highlighting the promising enhancement of hypergraph visualization for RAG.
\end{itemize}

\section{Hypergraph Visualization in RAG}
\label{sec:adv}
In this section, we clarify the situation of hypergraph visualization in RAG, including its 1) advantages, 2) feasibility, and 3) specific challenges.

\subsection{Advantages}
\label{sec:adv}
We first highlight why hypergraph visualization is a meaningful extension for hypergraph-based RAG.

\noindent\textbf{1) Structural Discrimination.}
Text-only retrieval often over-relies on lexical overlap and shallow semantics. In hypergraph-based RAG, many failure cases arise from \textit{structurally mismatched but lexically similar} evidence: candidate passages may share keywords with the query but encode different high-order relations among entities. Hypergraph visualization provides explicit spatial and grouping cues (e.g., overlap patterns, bridge entities, and cross-group links), helping MLLMs separate true relational matches from superficial textual matches. This enables finer-grained structural disambiguation before final answer generation.

\begin{note}
Hypergraph visualization enables RAG systems to achieve finer-grained structural discrimination in hypergraphs.
\end{note}

\noindent\textbf{2) Multi-hop Reasoning Composition.}
As the reasoning hop increases, relevant evidence becomes more scattered across entities, hyperedges, and chunks, making path composition fragile in text-only settings. Hypergraph visualization offers an externalized structural workspace: it exposes candidate evidence chains as visible substructures, making it easier for MLLMs to identify plausible multi-hop paths and reject shortcuts that break relational constraints. This is particularly important for deep reasoning, where correctness depends on composing multiple structurally consistent steps rather than selecting a single highly similar passage.

\begin{note}
Visual structural cues bring increasingly prominent benefits for complex deep multi-hop reasoning tasks.
\vspace{-5pt}
\end{note}

To empirically validate these advantages, we present experimental results in Sec.~\ref{sec:abl}.

\subsection{Feasibility}  
\label{sec:fea}
Apart from the benefits, we argue that hypergraph visualization is feasible in current RAG systems:

\noindent\textbf{(1) Usable MLLM Hypergraph Perception:}
Recent off-the-shelf MLLMs already demonstrate strong visual parsing ability for graph-like structures, including patterns like region overlap and bridge-like connectors (as evidenced in Sec.~\ref{sec:abl}). These capabilities are tightly aligned with the key signals required to understand the hypergraph knowledge structure in RAGs, such as identifying overlapping entities, analyzing multi-hop connections, and interpreting relational constraints encoded in hyperedges. Because current MLLMs can effectively parse hypergraph visualizations without fine-tuning, integrating visualization into RAG systems can be seamless.

\noindent\textbf{(2) Lightweight System Cost:} 
Although visualization introduces extra rendering and transfer overhead, this overhead is lightweight relative to end-to-end RAG latency, especially when retrieval and generation already dominate the runtime. A detailed efficiency comparison is provided in Sec.~\ref{sec:abl}, where the extra cost of visualization is very small compared to retrieval time. Therefore, hypergraph visualization does not introduce a significant computational burden, making it suitable for deployment in existing RAG frameworks.

\subsection{Specific Challenges}
\label{sec:cha}
Though beneficial and feasible for integrating hypergraph visualization into RAG, it still faces two non-trivial challenges that hinder its effectiveness.

\begin{figure}[t]
    \centering
    \vspace{-20pt}\includegraphics[width=0.8\linewidth]{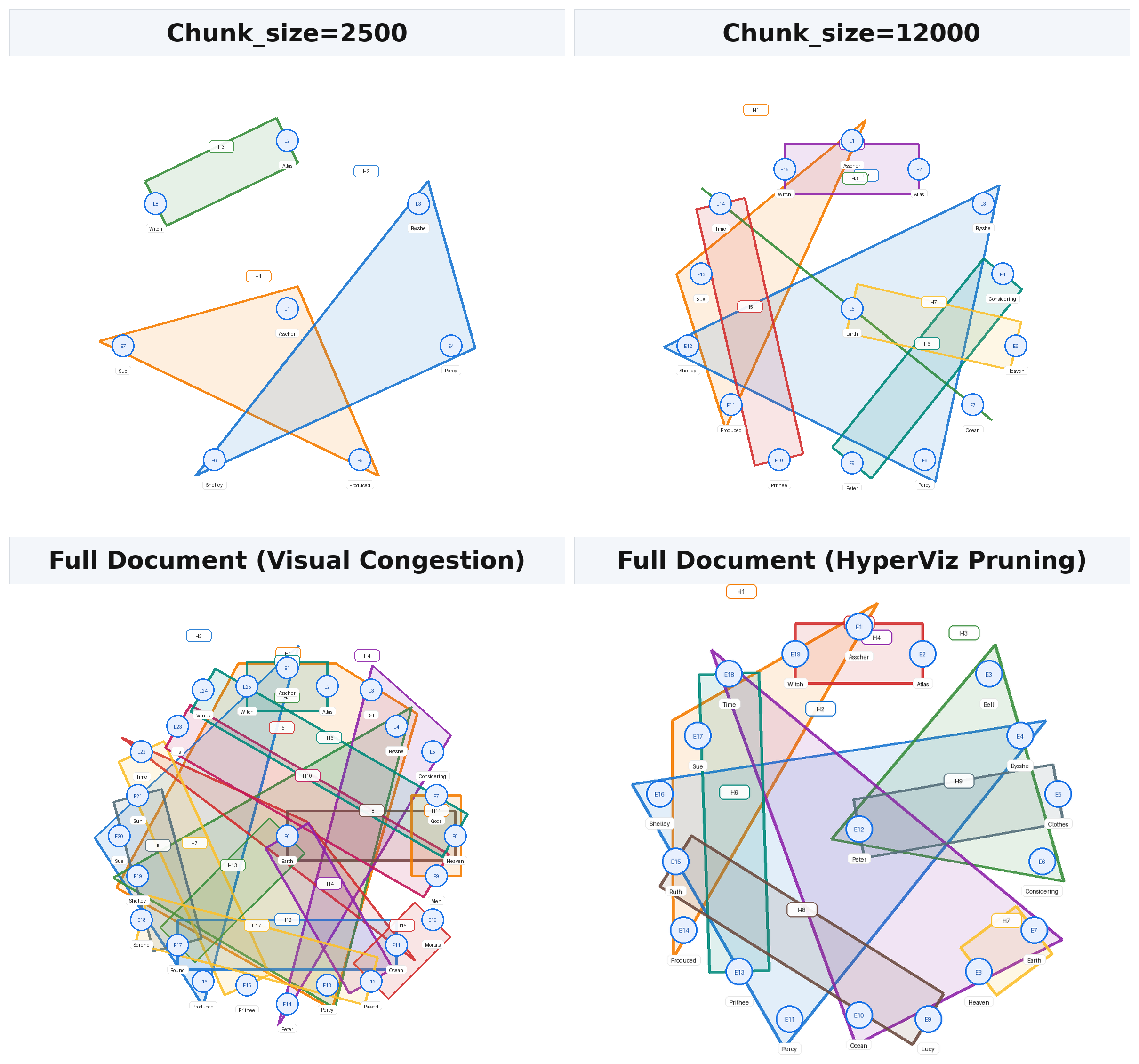}
    \caption{As the chunk size increases to even encompass the full document, more nodes and hyperedges are included in the visualization, leading to visual congestion that reduces structure clarity.}
    \label{fig:congestion}
    \vspace{-20pt}
\end{figure}

\paragraph{Visual Congestion.}  
Knowledge-intensive corpora often contain a high density of entities and multi-entity hyperedges, leading to visual clutter in hypergraph visualization. As shown in Fig.~\ref{fig:congestion}, the bigger chunk size results in more node aggregation, edge crossing, and structural overlap. Although these features reveal critical structural cues, excessive clutter can obscure reasoning paths and degrade performance. Dense node clusters may cause localized visual overload, whereas sparse low-hop structures risk leaving large blank spaces. Balancing information density and clarity requires employing reasonable sampling strategies and adaptive element size adjustments. To address these challenges, effective structure filtering mechanisms and spatial layout optimization techniques are essential for preserving key information and ensuring readability across varying graph densities.

\begin{figure}[t]
    \centering
    \vspace{-20pt}
    \includegraphics[width=\linewidth]{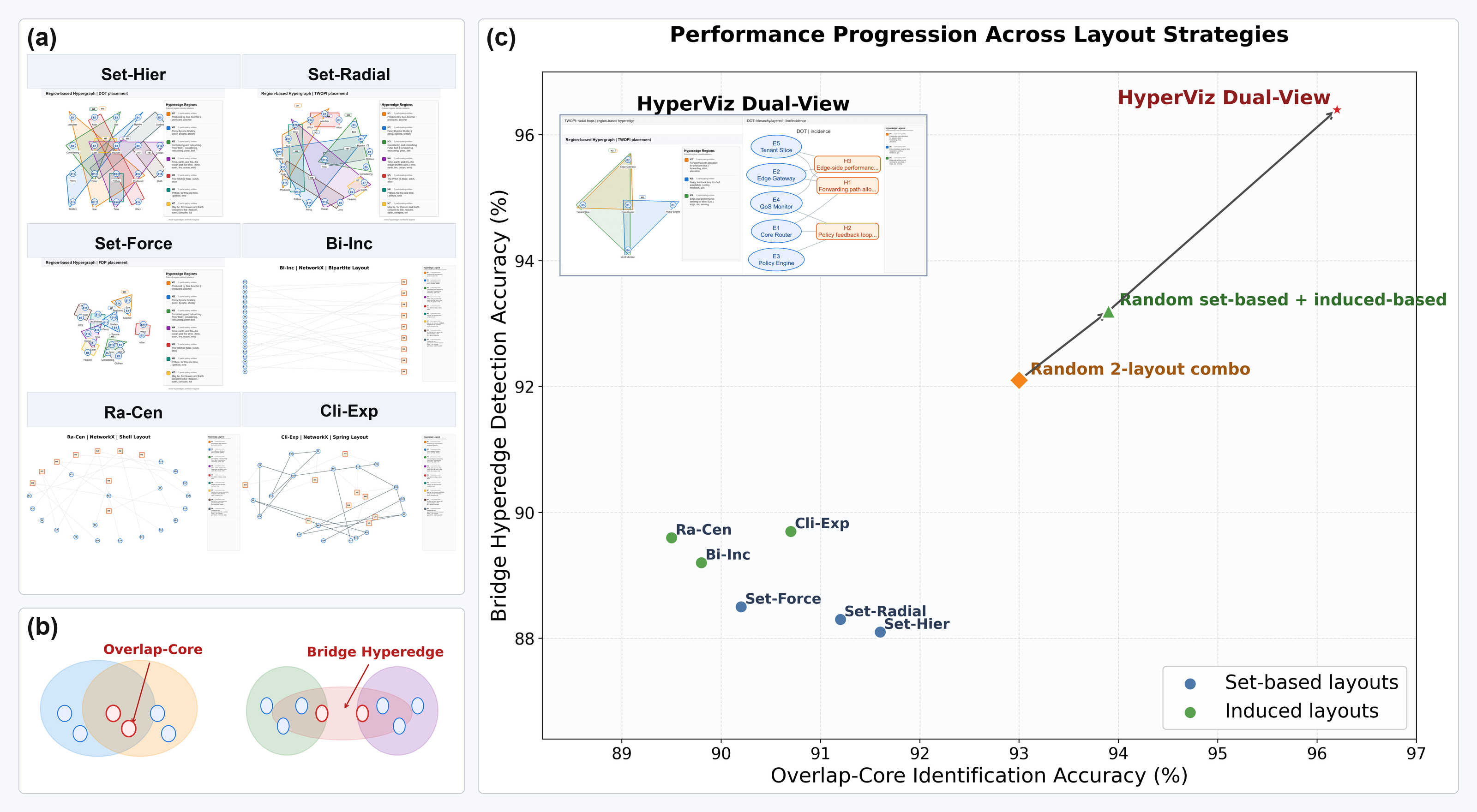}
    \vspace{-20pt}
    \caption{Rendering bias illustration. (a) Layout variants for the same hypergraph; (b) Structure perception tasks; (c) Performance gap caused by rendering bias.}
    \vspace{-10pt}
    \label{fig:bias}
\end{figure}

\paragraph{Rendering Bias.}
Another major challenge in hypergraph visualization is the uncontrollable rendering bias, where a single visualization emphasizes specific hypergraph properties, leading to partial or skewed structural understanding. Specifically, existing hypergraph layouts can be broadly categorized into two paradigms:  
i) \textit{Set-based layouts}, which represent high-order relationships (i.e., hyperedges) as set elements, such as through region containment, to highlight entity grouping relations.  
ii) \textit{Induced layouts}, which focus on direct belongingship between entities and their belonging groups by representing hyperedges as special nodes. In this approach, the nodes within each hyperedge are explicitly linked to the corresponding special node via pairwise edges.  
While both approaches provide valuable perspectives, they inherently emphasize different aspects of the hypergraph structure, which can lead to biased or incomplete interpretations depending on the chosen visualization method. To quantify this, we evaluate the Gemini-3 Flash model's structural perception on \textit{overlap-core identification} and \textit{bridge hyperedge detection}, where tasks are illustrated in Fig.~\ref{fig:bias}(b), with 2,000 hypergraphs derived from the UltraDomain Legal doc/chunks \cite{qian2025memorag}. As shown in Fig.~\ref{fig:bias}(c), significant discrepancies emerge between set-based and Induced visualizations. However, combining both layouts effectively mitigates unilateral perception deviations, with the strongest complementary effect observed when pairing set-based and induced views. This highlights the need for a unified paradigm in RAG systems to harness layout-induced biases for improved performance.

\section{HyperViz Toolkit}
\label{sec:hyperviz}
This section introduces \textbf{HyperViz}, a toolkit for hypergraph visualization in RAG systems. Taking the document hypergraph and text descriptions of entities/hyperedges, HyperViz generates visualization views that tackle visual congestion and rendering bias. By integrating structural and semantic considerations, it ensures document-based hypergraph visualization aligns with RAG requirements. To keep focus, we highlights its solutions to the aforementioned challenges in this section, with additional features are leaved in App.~\ref{app:chac}.

\subsection{Solution to Visual Congestion}
To address visual congestion, HyperViz employs a multi-step optimization strategy:

\noindent{\bf (1) Weighted Space Allocation:} Given a hypergraph $G = (V, E)$, where $V$ represents entity nodes and $E$ represents hyperedges, we assign distinct space weight coefficients $w_v$ for nodes and $w_e$ for hyperedges\footnote{$w_v, w_e \in [1,10]$ are grid-searched on large knowledge hypergraphs without data leakage under Gemini-3 Flash supervision via tasks in Fig.\ref{fig:bias}. More hyperparameter details are in App.~\ref{app:hyper}}. The total visual occupation cost is expressed as \( C = \sum_{v \in V} A_v \cdot w_v + \sum_{e \in E} L_e \cdot w_e\ \), where \( A_v \) and \( L_e \) denote the actual visual areas and lengths occupied by entity nodes and hyperedges, respectively. To ensure clarity, a maximum allowable cost \( C_{\text{max}} \) is defined as \( C_{\text{max}} = \gamma A \), where A is the area of the predefined canvas and  $\gamma$ controls the congestion threshold. Together, these formulations quantify the visual occupation by area and define an upper bound for maintaining clarity.

\noindent{\bf (2) Pruning Low-Importance Elements:} When $C$ > $C_{\text{max}}$, we rank all entities  in $G$ by an importance score $\text{Importance}(e)$, defined as 
$\text{Importance}(e) = \alpha\cdot\text{Centrality}(e) + \beta\cdot \text{Relevance}(e)$,
where $\text{Centrality}(e)$ is the degree centrality of the entity $e$ in $G$, and $\text{Relevance}(e)$ reflects a query-specific semantic alignment by the BM25 score between the text descriptions of $e$ and Query $Q$, and $\alpha$, $\beta$ denote weighting coefficients. Based on this ranking, pruning is applied to the low-importance nodes, and they are eliminated from their associated hyperedges until $C \leq C_{\text{max}}$. This ensures that the resulting pruned hypergraph retains critical high-order relationships while significantly reducing visual clutter, improving interpretability and usability.

\noindent{\bf (3) Global Uniform Scaling}: The filtered hypergraph is then scaled to fit the canvas dimensions, maximizing canvas utilization and eliminating marginal blank spaces to avoid space wasting. 

In summary, the multi-step pipeline effectively mitigates visual congestion. As an illustration shown in Fig.~\ref{fig:congestion}, the pruned visualization generated by HyperViz \textit{preserves clarity} while \textit{retaining key structural and semantic information} in the document knowledge.

\subsection{Solution to Rendering Bias}

To fit diverse scenarios and increase the usability of Hyperviz, we categorize the addressing of rendering bias into two scenarios:  

\noindent{\bf (1) User-driven Customization:}  
To support tasks requiring specific rendering biases, HyperViz employs a layout-centric design encompassing both set-based and Induced paradigms, each offering three layout variants. The set-based layouts include i) Set-Hier, ii) Set-Radial, and iii) Set-Force, while the Induced layouts consist of i) Bi-Inc, ii) Ra-Cen, and iii) Cli-Exp. They are illustrated in Fig.~\ref{fig:bias}(a), with detailed descriptions provided in App.~\ref{app:intro_layout}. Users can flexibly select a single layout or combine multiple layouts as needed.

\noindent{\bf (2) Automated Bias Mitigation:}  
To address uncontrolled rendering biases, HyperViz incorporates a \textbf{probe-driven dual-view} mechanism as the default approach. Specifically, a small batch of $k$ queries ($k=15$ in our experiments, random from training sets) is used to evaluate all nine combinations of set-based and induced-graph layouts, inspired by their complementary strengths.  
The layout combination that best enhances hypergraph perception on \textit{overlap-core identification} and \textit{bridge hyperedges detection} (illustrated in Fig.~\ref{fig:bias}(b))\footnote{Both tasks are chosen by their similarity to essential RAG requirements, such as identifying key entities within groups and uncovering critical relationships among entity clusters.} is selected, and the two selected layouts are rendered side-by-side in a dual-view format within a single image, as demonstrated in Fig.~\ref{fig:bias}(c) left-top. Results in Fig.~\ref{fig:bias}(c) show that the dynamic probe-driven dual-view significantly improves hypergraph perception on both tasks compared to either single layouts or their random combinations. This finally benefits the RAG quality (Sec.~\ref{sec:abl}).

\subsection{Plug-and-Play Integration}
HyperViz serves as a middleware that bridges knowledge hypergraphs and visualization workflows tailored to RAG scenarios. It seamlessly integrates with existing RAG pipelines in a plug-and-play manner, offering well-encapsulated interfaces to ensure ease of use for users.

\begin{figure*}[t]
    \centering
    \includegraphics[width=\linewidth]{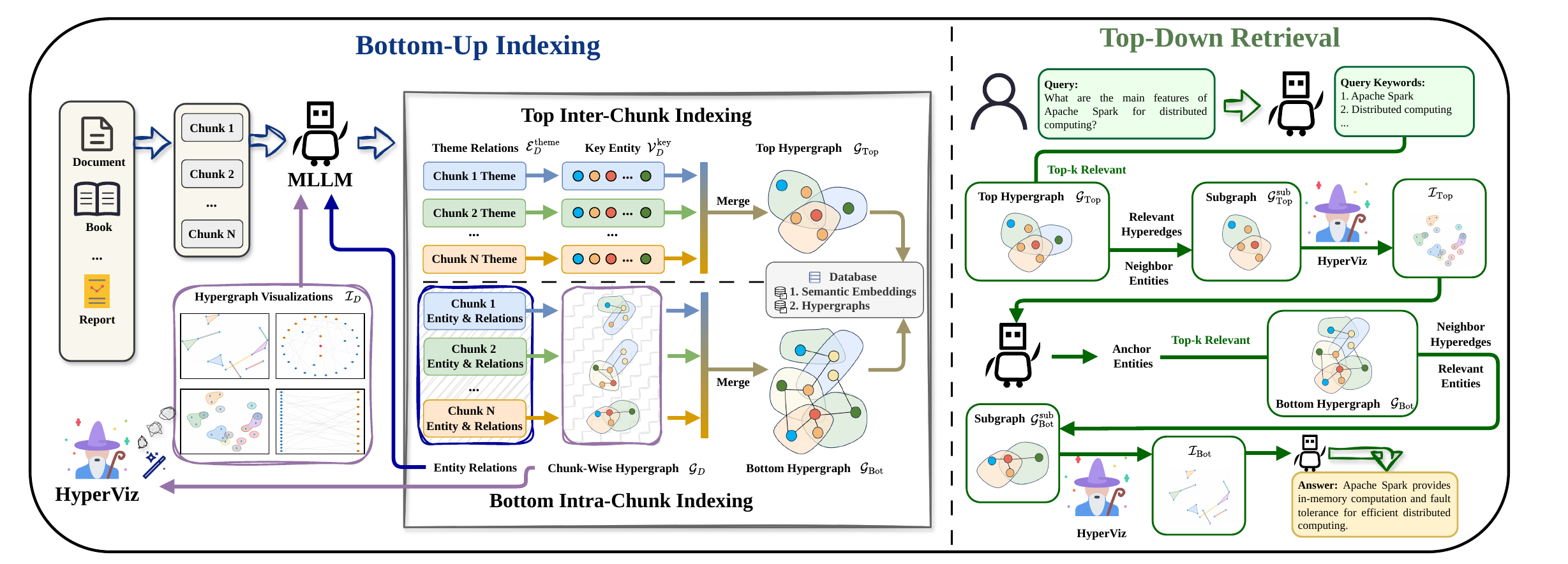}
    \vspace{-16pt}
    \caption{The overall framework of VizRAG.}
    \label{fig:framework}
    \vspace{-15pt}
\end{figure*}

\section{VizRAG Framework}
\label{sec:vizrag}
In this section, we introduce the VizRAG framework (illustrated in Fig.~\ref{fig:framework}), where visualization plays a dual role in both indexing and retrieving.

\subsection{Vision-awared Bottom-up Indexing}
Given a corpus $\mathcal{D}$ (e.g., books, reports, or manuals), we segment it into a set of overlapping chunks using a fixed-length sliding window to preserve semantic continuity:
\begin{equation*}
    \mathcal{D} = \{D_1, D_2, \dots, D_N\},
\end{equation*}
where $D_i$ represents the $i$-th chunk, serving as the fundamental unit for subsequent processing.

We employ a two-layer hierarchical hypergraph indexing approach:  
(i) bottom-layer intra-chunk indexing, capturing fine-grained intra-chunk semantics.  
(ii) top-layer inter-chunk indexing, capturing global relationships across chunks.

\subsubsection{Bottom Intra-Chunk Indexing}
For each chunk $D_i$, we extract entities (e.g., persons, events, organizations) and their descriptions using LLMs. These entities form the vertex set $\mathcal{V}_{D_i}$ of the fine-grained entity hypergraph. Semantic relationships among the entities are used to construct hyperedges $\mathcal{E}_{D_i}$, capturing high-order associations such as co-occurrence in events or causal links:
\begin{equation*}
\left\{
\begin{aligned}
    &\mathcal{V}_{D_i} = \mathrm{LLM}(\mathcal{P}_\text{ext\_entity}(D_i)), \\
    &\mathcal{E}_{D_i} = \mathrm{LLM}(\mathcal{P}_\text{ext\_relation}(D_i, \mathcal{V}_{D_i})),
\end{aligned}
\right.  
\end{equation*}
where $\mathcal{P}_\text{ext\_entity}$ and $\mathcal{P}_\text{ext\_relation}$ are prompts designed for entity and hyperedge extraction (detailed in Appendix~\ref{app:prompt}).  

Once the hypergraph $\mathcal{G}_{D_i} = \{\mathcal{V}_{D_i}, \mathcal{E}_{D_i}\}$ is constructed, we generate its hypergraph visualization using HyperViz:
\begin{equation*}
\mathcal{I}_{D_i} = \text{HyperViz}(\mathcal{G}_{D_i}).
\end{equation*}

\subsubsection{Top Inter-Chunk Indexing}
Using predefined prompts $\mathcal{P}_\text{ext\_theme}$ and $\mathcal{P}_\text{ext\_key}$ (in App.~\ref{app:prompt}), MLLMs are guided to perform semantic parsing on each chunk $D_i$, extracting its theme $\mathcal{E}^\text{theme}_{D_i}$ and a subset of key entities $\mathcal{V}^\text{key}_{D_i}$ relevant to the theme. Importantly, the intra-chunk knowledge structure visualized in $\mathcal{I}_{D_i}$ is integrated into the MLLM to enhance semantic understanding:
\begin{equation*}
\left\{
\begin{aligned}
    &\mathcal{E}^\text{theme}_{D_i} = \mathrm{MLLM}(\mathcal{P}_\text{ext\_theme}(D_i, \mathcal{I}_{D_i})), \\
    &\mathcal{V}^\text{key}_{D_i} = \mathrm{MLLM}(\mathcal{P}_\text{ext\_key}(\mathcal{E}^\text{theme}_{D_i}, D_i, \mathcal{I}_{D_i}, \mathcal{V}_{D_i})).
\end{aligned}
\right.
\end{equation*}

\subsubsection{Hierarchical Hypergraph Merging}
Merging all extracted entities, their relations, and chunk themes from all chunks, we get a fine-grained bottom hypergraph $\mathcal{G_\text{Bot}}$ and a theme-aware top hypergraph $\mathcal{G_\text{top}}$, which are connected via shared vertices (i.e., theme-related key entities \{$V_\text{key}$\}). Both hypergraphs are stored in a hypergraph database.
\begin{equation*}
\left\{
\begin{aligned}
    &\mathcal{G_\text{Bot}} = \{\textstyle \bigcup_{D_i\in \mathcal{D}}\mathcal{V}_{D_i},\textstyle \bigcup_{D_i\in \mathcal{D}}\mathcal{E}_{D_{i}}\}, \\
    &\mathcal{G_\text{Top}} = \{\textstyle \bigcup_{D_i\in \mathcal{D}}\mathcal{V}^\text{key}_{D_i},\textstyle \bigcup_{D_i\in \mathcal{D}}\mathcal{E}^\text{theme}_{D_{i}}\}.
\end{aligned}
\right.
\end{equation*}

\subsection{Vision-enhanced Top-down Retrieval}
VizRAG employs a visualization-guided top-down retrieval strategy, leveraging rendered hypergraph images as structural bridges from the global theme layer $\mathcal{G}_\text{Top}$ to the fine-grained entity layer $\mathcal{G}_\text{Bot}$. Given a query $q$, the keywords are first extracted:
\begin{equation*}
\mathcal{X}_\text{query} = \mathrm{LLM}(\mathcal{P}_\text{keyword}(q)),
\end{equation*}
where $\mathcal{X}_\ast=\{x_1,x_2,\ldots\}$ and $\mathcal{P}_\text{keyword}$ is in App.~\ref{app:prompt}.

\subsubsection{Top Retrieval and Locating}
These keywords $\mathcal{X}_\text{query}$ are used to retrieve the top-$k$ relevant hyperedges $\mathcal{E}_{rel}$ from $\mathcal{G}_\text{Top}$ based on vector similarity $\mathcal{R}(\cdot, \cdot)$, where $\mathcal{E}^\text{Top}$ are edges in $\mathcal{G}_\text{Top}$:
\begin{equation*}
\mathcal{E}_\text{rel} = \textstyle \bigcup_{x_i \in \mathcal{X}_\text{query}} \mathcal{R}(x_i,\, \mathcal{E}^\text{Top}).
\end{equation*}
We then construct a h-hop induced subgraph $\mathcal{G}_\text{Top}^\text{sub}$ centered around the retrieved  hyperedges $\mathcal{E}_\text{rel}$ and visualize it using HyperViz:
\begin{equation*}
\mathcal{I}_\text{Top} = \text{HyperViz}(\text{Subgraph}(\mathcal{G}_\text{Top}, \mathcal{E}_\text{rel}, \text{hops}=h)).
\end{equation*}
VizRAG then exploits the MLLM's visual structural perception to locate key entities from $\mathcal{I}_\text{Top}$, union with the direct neighbors of hyperedges $\mathcal{E}_\text{rel}$:
\begin{align*}
\hat{\mathcal{V}} =& \mathrm{MLLM}(\mathcal{P}_\text{locate}(q,\, \mathcal{I}_\text{Top},\, \mathcal{C}_\text{e\_rel})) \cup \mathcal{V}_{dif}, \\
\mathcal{V}_{dif} =& \textstyle \bigcup_{e_i \in {\mathcal{E}_{rel}}} \mathcal{N}(e_i,\, \mathcal{G}_\text{Top}),
\end{align*}
where $\mathcal{N}$ denotes retrieving neighbors from the hypergraph, $\mathcal{C}_\text{e\_rel}$ is the relevant context of $\text{e\_rel} \in \mathcal{E}_\text{rel}$, $\mathcal{P}_\text{locate}$  prompts the MLLM to locate query-relevant entities in the rendered subgraph structure.

\subsubsection{Bottom Retrieval and Answer Generation}
With $\hat{\mathcal{V}}$ as entity anchors, we retrieve the top-$k$ matching vertices from $\mathcal{G}_\text{Bot}$ and expand context via their neighbor hyperedges, where $\mathcal{V}_\text{Bot} = \{\mathcal{V}_{D_i}\}$:
\begin{align*}
\mathcal{V}_\text{rel} &= \textstyle \bigcup_{v_i \in \hat{\mathcal{V}}} \mathcal{R}(v_i,\mathcal{V}_\text{Bot}),\\
\mathcal{E}_\text{dif} &= \textstyle \bigcup_{v_i \in \mathcal{V}_\text{rel}} \mathcal{N}(v_i,\mathcal{G}_\text{Bot}).
\end{align*}
The retrieved entity subgraph is rendered as a visual cue $\mathcal{I}_\text{Bot}$ to expose fine-grained evidence paths:
\begin{equation*}
\mathcal{I}_\text{Bot} = \text{HyperViz}(\text{Subgraph}(\mathcal{G}_\text{Bot}, \mathcal{V}_\text{rel}, \text{hops}=h)).
\end{equation*}
Finally, $\mathcal{V}_\text{rel}$, $\mathcal{E}_\text{dif}$, their corresponding contexts $\mathcal{C}_\text{v\_rel}, \mathcal{C}_\text{e\_dif}$, and the visualization $\mathcal{I}_\text{Bot}$ are jointly fed to the MLLM to generate the final answer:
\begin{equation*}
\mathcal{A} = \mathrm{MLLM}(q,\, \mathcal{I}_\text{Bot},\, \mathcal{V}_\text{rel},\, \mathcal{E}_\text{dif},\, \mathcal{C}_\text{v\_rel},\, \mathcal{C}_\text{e\_dif}).
\end{equation*}

\section{Experiments}
In this section, we perform extensive experiments to evaluate the proposed methods. The experimental setup is aligned with prior works \cite{hu2026cog, feng2025hyperrag}. 

\begin{table*}[ht]
    \centering
    \vspace{-10pt}
    \scriptsize
    \setlength{\tabcolsep}{1mm}
    \small
    \begin{tabular}{lcccccccccc}
        \toprule
        & \multicolumn{2}{c}{\textbf{Mix}} & \multicolumn{2}{c}{\textbf{CS}} & \multicolumn{2}{c}{\textbf{Agriculture}} & \multicolumn{2}{c}{\textbf{Neurology}} & \multicolumn{2}{c}{\textbf{Pathology}} \\
        \midrule
        & GraphRAG & \textbf{VizRAG} & GraphRAG & \textbf{VizRAG} & GraphRAG & \textbf{VizRAG} & GraphRAG & \textbf{VizRAG} & GraphRAG & \textbf{VizRAG} \\
        \cmidrule(lr){2-3} \cmidrule(lr){4-5} \cmidrule(lr){6-7} \cmidrule(lr){8-9} \cmidrule(lr){10-11}
        Comp. & 28.7\% & \textbf{71.3\%} & 24.5\% & \textbf{75.5\%} & 20.8\% & \textbf{79.2\%} & 22.7\% & \textbf{77.3\%} & 20.5\% & \textbf{79.5\%} \\
        Empo. & 24.6\% & \textbf{75.4\%} & 23.2\% & \textbf{76.8\%} & 14.8\% & \textbf{85.2\%} & 15.6\% & \textbf{84.4\%} & 11.7\% & \textbf{88.3\%} \\
        Rele. & 33.9\% & \textbf{66.1\%} & 27.5\% & \textbf{72.5\%} & 23.1\% & \textbf{76.9\%} & 25.2\% & \textbf{74.8\%} & 19.6\% & \textbf{80.4\%} \\
        Cons. & 28.9\% & \textbf{71.1\%} & 23.7\% & \textbf{76.3\%} & 17.9\% & \textbf{82.1\%} & 19.5\% & \textbf{80.5\%} & 18.6\% & \textbf{81.4\%} \\
        Clar. & 34.7\% & \textbf{65.3\%} & 24.8\% & \textbf{75.2\%} & 26.5\% & \textbf{73.5\%} & 24.2\% & \textbf{75.8\%} & 18.5\% & \textbf{81.5\%} \\
        Logi. & 27.8\% & \textbf{72.2\%} & 25.1\% & \textbf{74.9\%} & 15.9\% & \textbf{84.1\%} & 21.6\% & \textbf{78.4\%} & 17.8\% & \textbf{82.2\%} \\
        Overall & 29.8\% & \textbf{70.2\%} & 24.8\% & \textbf{75.2\%} & 19.8\% & \textbf{80.2\%} & 21.5\% & \textbf{78.5\%} & 17.8\% & \textbf{82.2\%} \\
        \midrule
        & LightRAG & \textbf{VizRAG} & LightRAG & \textbf{VizRAG} & LightRAG & \textbf{VizRAG} & LightRAG & \textbf{VizRAG} & LightRAG & \textbf{VizRAG} \\
        \cmidrule(lr){2-3} \cmidrule(lr){4-5} \cmidrule(lr){6-7} \cmidrule(lr){8-9} \cmidrule(lr){10-11}
        Comp. & 30.5\% & \textbf{69.5\%} & 22.8\% & \textbf{77.2\%} & 16.6\% & \textbf{83.4\%} & 20.9\% & \textbf{79.1\%} & 22.9\% & \textbf{77.1\%} \\
        Empo. & 22.2\% & \textbf{77.8\%} & 19.5\% & \textbf{80.5\%} & 12.6\% & \textbf{87.4\%} & 15.7\% & \textbf{84.3\%} & 17.6\% & \textbf{82.4\%} \\
        Rele. & 28.9\% & \textbf{71.1\%} & 19.8\% & \textbf{80.2\%} & 17.7\% & \textbf{82.3\%} & 20.7\% & \textbf{79.3\%} & 24.8\% & \textbf{75.2\%} \\
        Cons. & 26.8\% & \textbf{73.2\%} & 21.5\% & \textbf{78.5\%} & 13.7\% & \textbf{86.3\%} & 17.8\% & \textbf{82.2\%} & 19.7\% & \textbf{80.3\%} \\
        Clar. & 30.6\% & \textbf{69.4\%} & 16.7\% & \textbf{83.3\%} & 14.8\% & \textbf{85.2\%} & 18.5\% & \textbf{81.5\%} & 18.8\% & \textbf{81.2\%} \\
        Logi. & 27.7\% & \textbf{72.3\%} & 21.6\% & \textbf{78.4\%} & 15.6\% & \textbf{84.4\%} & 18.8\% & \textbf{81.2\%} & 18.9\% & \textbf{81.1\%} \\
        Overall & 27.8\% & \textbf{72.2\%} & 20.3\% & \textbf{79.7\%} & 15.2\% & \textbf{84.8\%} & 18.7\% & \textbf{81.3\%} & 20.4\% & \textbf{79.6\%} \\
        \midrule
        & HiRAG & \textbf{VizRAG} & HiRAG & \textbf{VizRAG} & HiRAG & \textbf{VizRAG} & HiRAG & \textbf{VizRAG} & HiRAG & \textbf{VizRAG} \\
        \cmidrule(lr){2-3} \cmidrule(lr){4-5} \cmidrule(lr){6-7} \cmidrule(lr){8-9} \cmidrule(lr){10-11}
        Comp. & 36.6\% & \textbf{63.4\%} & 32.8\% & \textbf{67.2\%} & 33.7\% & \textbf{66.3\%} & 27.5\% & \textbf{72.5\%} & 32.9\% & \textbf{67.1\%} \\
        Empo. & 31.5\% & \textbf{68.5\%} & 28.7\% & \textbf{71.3\%} & 28.5\% & \textbf{71.5\%} & 23.6\% & \textbf{76.4\%} & 29.6\% & \textbf{70.4\%} \\
        Rele. & 37.7\% & \textbf{62.3\%} & 39.8\% & \textbf{60.2\%} & 36.7\% & \textbf{63.3\%} & 27.6\% & \textbf{72.4\%} & 33.8\% & \textbf{66.2\%} \\
        Cons. & 31.6\% & \textbf{68.4\%} & 32.6\% & \textbf{67.4\%} & 29.6\% & \textbf{70.4\%} & 24.7\% & \textbf{75.3\%} & 29.9\% & \textbf{70.1\%} \\
        Clar. & 38.6\% & \textbf{61.4\%} & 42.7\% & \textbf{57.3\%} & 36.8\% & \textbf{63.2\%} & 23.5\% & \textbf{76.5\%} & 32.7\% & \textbf{67.3\%} \\
        Logi. & 32.6\% & \textbf{67.4\%} & 32.7\% & \textbf{67.3\%} & 30.8\% & \textbf{69.2\%} & 23.8\% & \textbf{76.2\%} & 28.6\% & \textbf{71.4\%} \\
        Overall & 34.8\% & \textbf{65.2\%} & 34.9\% & \textbf{65.1\%} & 32.7\% & \textbf{67.3\%} & 25.1\% & \textbf{74.9\%} & 31.2\% & \textbf{68.8\%} \\
        \midrule
        & Hyper-RAG & \textbf{VizRAG} & Hyper-RAG & \textbf{VizRAG} & Hyper-RAG & \textbf{VizRAG} & Hyper-RAG & \textbf{VizRAG} & Hyper-RAG & \textbf{VizRAG} \\
        \cmidrule(lr){2-3} \cmidrule(lr){4-5} \cmidrule(lr){6-7} \cmidrule(lr){8-9} \cmidrule(lr){10-11}
        Comp. & 37.7\% & \textbf{62.3\%} & 38.6\% & \textbf{61.4\%} & 40.8\% & \textbf{59.2\%} & 31.6\% & \textbf{68.4\%} & 33.8\% & \textbf{66.2\%} \\
        Empo. & 34.9\% & \textbf{65.1\%} & 35.7\% & \textbf{64.3\%} & 31.7\% & \textbf{68.3\%} & 28.6\% & \textbf{71.4\%} & 28.8\% & \textbf{71.2\%} \\
        Rele. & 47.2\% & \textbf{52.8\%} & 40.7\% & \textbf{59.3\%} & 38.7\% & \textbf{61.3\%} & 39.2\% & \textbf{60.8\%} & 28.7\% & \textbf{71.3\%} \\
        Cons. & 34.6\% & \textbf{65.4\%} & 36.2\% & \textbf{63.8\%} & 36.8\% & \textbf{63.2\%} & 29.8\% & \textbf{70.2\%} & 26.6\% & \textbf{73.4\%} \\
        Clar. & 46.6\% & \textbf{53.4\%} & 39.7\% & \textbf{60.3\%} & 33.6\% & \textbf{66.4\%} & 32.7\% & \textbf{67.3\%} & 24.7\% & \textbf{75.3\%} \\
        Logi. & 35.6\% & \textbf{64.4\%} & 37.6\% & \textbf{62.4\%} & 34.7\% & \textbf{65.3\%} & 26.8\% & \textbf{73.2\%} & 28.7\% & \textbf{71.3\%} \\
        Overall & 39.4\% & \textbf{60.6\%} & 38.1\% & \textbf{61.9\%} & 36.0\% & \textbf{64.0\%} & 31.4\% & \textbf{68.6\%} & 28.5\% & \textbf{71.5\%} \\
        \midrule
        & Cog-RAG & \textbf{VizRAG} & Cog-RAG & \textbf{VizRAG} & Cog-RAG & \textbf{VizRAG} & Cog-RAG & \textbf{VizRAG} & Cog-RAG & \textbf{VizRAG} \\
        \cmidrule(lr){2-3} \cmidrule(lr){4-5} \cmidrule(lr){6-7} \cmidrule(lr){8-9} \cmidrule(lr){10-11}
        Comp. & 40.7\% & \textbf{59.3\%} & 42.6\% & \textbf{57.4\%} & 46.6\% & \textbf{53.4\%} & 37.7\% & \textbf{62.3\%} & 39.7\% & \textbf{60.3\%} \\
        Empo. & 39.5\% & \textbf{60.5\%} & 40.5\% & \textbf{59.5\%} & 37.8\% & \textbf{62.2\%} & 34.6\% & \textbf{65.4\%} & 34.7\% & \textbf{65.3\%} \\
        Rele. & 44.7\% & \textbf{55.3\%} & 44.6\% & \textbf{55.4\%} & 42.5\% & \textbf{57.5\%} & 43.7\% & \textbf{56.3\%} & 34.7\% & \textbf{65.3\%} \\
        Cons. & 40.6\% & \textbf{59.4\%} & 41.6\% & \textbf{58.4\%} & 41.7\% & \textbf{58.3\%} & 35.8\% & \textbf{64.2\%} & 33.5\% & \textbf{66.5\%} \\
        Clar. & 47.6\% & \textbf{52.4\%} & 45.7\% & \textbf{54.3\%} & 39.8\% & \textbf{60.2\%} & 38.7\% & \textbf{61.3\%} & 29.8\% & \textbf{70.2\%} \\
        Logi. & 41.6\% & \textbf{58.4\%} & 43.7\% & \textbf{56.3\%} & 40.7\% & \textbf{59.3\%} & 32.6\% & \textbf{67.4\%} & 34.7\% & \textbf{65.3\%} \\
        Overall & 42.4\% & \textbf{57.6\%} & 43.1\% & \textbf{56.9\%} & 41.5\% & \textbf{58.5\%} & 37.2\% & \textbf{62.8\%} & 34.5\% & \textbf{65.5\%} \\
        \bottomrule
    \end{tabular}
    \vspace{-10pt}
    \caption{Average win rates of six evaluation metrics across five datasets. The comparison is made between baselines and VizRAG. Among them, we refer to the six metrics as Comp. (Comprehensiveness), Empo. (Empowerment), Rele. (Relevance), Cons. (Consistency), Clar. (Clarity), and Logi. (Logical).}
    \vspace{-15pt}
    \label{tab1}
\end{table*}
\subsection{Experimental Setup}
To systematically evaluate our method across diverse application scenarios, we adopt five datasets from two benchmarks: Mix, CS, and Agriculture from the UltraDomain benchmark \cite{qian2025memorag}, and Neurology and Pathology from the MIRAGE benchmark \cite{xiong2024benchmarking}. UltraDomain covers typical RAG applications across different domains, while MIRAGE focuses on medical question answering and domain-specific knowledge coverage. Data statistics are in App.~\ref{app:data}.

Based on domain consistency and semantic correlation within the texts, we categorize the datasets into three types to enable a comprehensive analysis of the model's adaptability: 
\textbf{\textit{Cross-domain Sparse}} (Mix): Fragmented passages from unrelated domains with weak semantic coherence.
\textbf{\textit{Intra-domain Sparse}} (CS, Agriculture): Domain-specific documents with weak inter-passage context.
\textbf{\textit{Intra-domain Dense}} (Neurology, Pathology): Highly structured medical textbooks with strong semantic continuity from MIRAGE. 
Additionally, we follow the data processing and query procedure of LightRAG, utilizing Gemini-3 Flash to generate complex, document-related queries. Besides, to enhance the robustness and reduce stochasticity, we evaluate 10 times for each result. All experiments are conducted on 4 A100-80G GPUs.

\begin{figure*}[t]
    \centering
    \vspace{-21pt}
    \includegraphics[width=\linewidth]{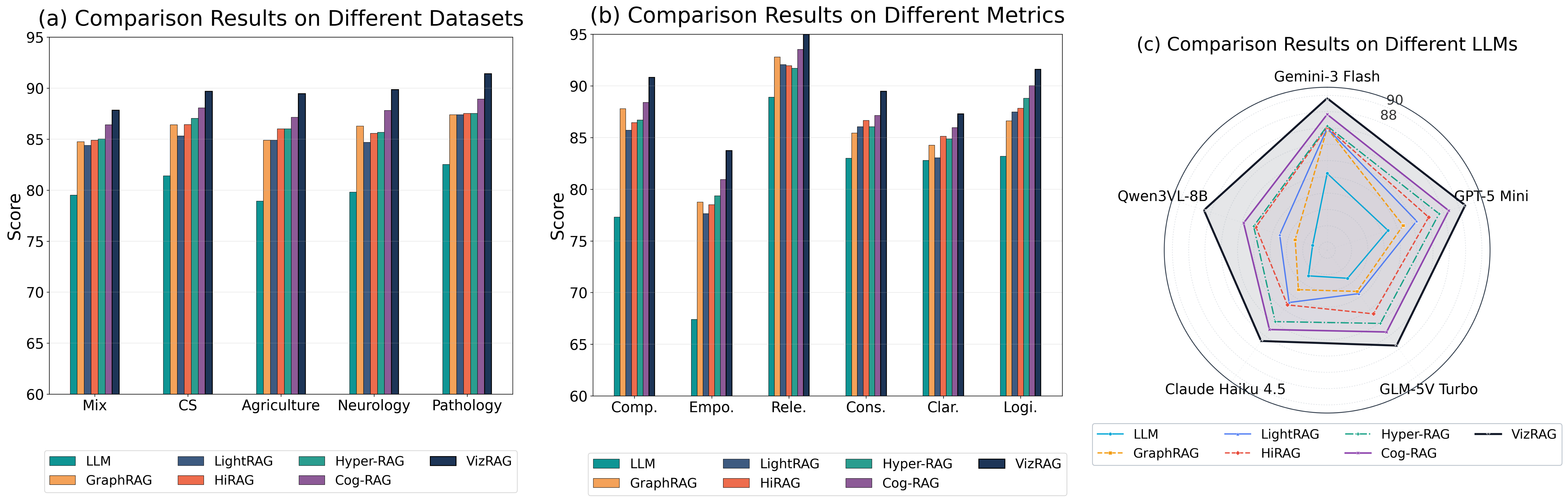}
    \vspace{-25pt}
    \caption{Test results by scoring. (a) is the comparison results on five datasets; (b) is the comparison results on six
dimensions; (c) shows the evaluation results on different LLMs.
}
    \label{fig:fig3}
    \vspace{-15pt}
\end{figure*}

\noindent{\bf Baselines.}
We compared our approach with the state-of-the-art and popular Graph-enhanced RAG methods: GraphRAG \cite{edge2024local}, LightRAG \cite{guo2024lightrag}, HiRAG \cite{huang2025retrieval}, and hypergraph RAGs including Hyper-RAG \cite{feng2025hyper} and Cog-RAG \cite{hu2026cog}. The baseline details are in App.~\ref{app:baseline}.

\noindent{\bf Implementation Details.}
To ensure fairness and consistency for both the baselines and proposed methods, we validate on five different LLMs for information extraction and question answering, including Gemini-3 Flash \cite{gemini3}, GPT-5 Mini \cite{gpt5}, GLM-5V Turbo \cite{glm5v}, Claude Haiku 4.5 \cite{claude}, and Qwen3VL-8B \cite{Qwen3-VL}. The result evaluation is default on Gemini-3 Flash, as well as the text-embedding-3-small embedding model for vector encoding and retrieval tasks.

\noindent{\bf Evaluation Metrics.}
Following the recent works, we adopt two evaluation strategies: Selection-based \cite{guo2024lightrag, huang2025retrieval} and Score-based \cite{wang2024leave, feng2025hyper}, providing both relative and absolute perspectives on model performance.
\textbf{\textit{The Selection-based}} evaluation uses LLMs to report win rates of answer quality between two methods. 
\textbf{\textit{The Score-based}} evaluation employs LLMs to score responses of different methods. Both strategies assess models from six dimensions: Comprehensiveness, Empowerment, Relevance, Consistency, Clarity, and Logical. We report both per-dimension and overall average scores. More Evaluation Details are in App.~\ref{app:eval}.

\subsection{Main Results}
Our primary results are presented in Tab.~\ref{tab1} and Fig.~\ref{fig:fig3}, and more results are provided in App.~\ref{app:more_results}. VizRAG consistently outperforms all baselines across multiple dimensions. Additionally, we have several key insights:

1) \textbf{VizRAG consistently wins against all baselines on selection-based evaluation.}
As shown in Tab.~\ref{tab1}, the overall win rates are 70.2\%--82.2\% vs. GraphRAG, 72.2\%--84.8\% vs. LightRAG, and 65.1\%--74.9\% vs. HiRAG across five datasets. VizRAG also outperforms strong hypergraph baselines, with 60.6\%--71.5\% vs. Hyper-RAG and 56.9\%--65.5\% vs. Cog-RAG. This confirms the effectiveness of VizRAG on pairwise comparison.

2) \textbf{VizRAG brings robust and comprehensive improvement on response quality.}
Figs.~\ref{fig:fig3}(a) and \ref{fig:fig3}(b) show that VizRAG achieves higher scores across \textit{all datasets} and \textit{all evaluation dimensions}, and Fig.~\ref{fig:fig3}(c) shows the same trend across \textit{all tested LLM backbones}. These results verify that adding visual hypergraph cues delivers persistent and robust gains for hypergraph-based RAG systems.

\subsection{Ablation Study} 
\noindent{\bf Effects of Method Components.}
\label{sec:abl} We report score-based ablations on three representative datasets in Tab.~\ref{tab2} to validate the necessity of key components in our proposed methods: 1) visualization in indexing, 2) visualization in retrieval, 3) HyperViz for addressing visual congestion, and 4)  HyperViz for addressing rendering bias. As the results shown in Tab.~\ref{tab2}, removing any component leads to obvious performance degradation on all datasets, confirming their empirical effectiveness. 

\begin{table}[ht]
    \centering
    \vspace{-5pt}
    \scriptsize
    \setlength{\tabcolsep}{2.8pt}
    \begin{tabular}{lccc}
    \toprule
    \textbf{Models} & \textbf{Mix} & \textbf{CS} & \textbf{Neurology} \\
    \midrule
    \textsc{VizRAG} & 88.0 & 89.7 & 89.9 \\
    \textit{w./o. Indexing Visualization} & 87.3 & 88.5 & 89.4 \\
    \textit{w./o. Retrieval Visualization} & 86.8 & 87.3 & 88.2 \\
    \textit{w./o. HyperViz (Address Visual Congestion)} & 87.5 & 88.1 & 88.9 \\
    \textit{w./o. HyperViz (Probe-driven Dual-View)} & 87.9 & 88.4 & 89.2 \\
    \bottomrule
    \end{tabular}
    \vspace{-6pt}
    \caption{Ablation results by overall scores. All ablated variants underperform full VizRAG on all datasets.}
    \label{tab2}
    \vspace{-15pt}
\end{table}

\noindent{\bf Advantages of Structural Discrimination.} To evaluate whether hypergraph visualization improves structural disambiguation, we define a binary classification task: distinguishing between structurally aligned positive candidates and lexically similar but structurally mismatched negative candidates. Using 200 query groups from the UltraDomain dataset \cite{qian2025memorag}, each query is paired with one correct match (positive) and one distractor (negative). The goal is to classify each candidate as ``positive'' or ``negative''. As shown in Tab.~\ref{tab:adv_disambiguation}, integrating hypergraph visualization into the Hypergraph RAG pipeline \cite{feng2025hyperrag} improves accuracy from 71.5\% to 79.0\%, highlighting its effectiveness in resolving structural ambiguities.

\begin{table}[ht]
\centering
\resizebox{0.95\linewidth}{!}{
\begin{tabular}{lc}
\hline
Method & Accuracy (\%)\\
\hline
Hypergraph RAG w/o Visualization & 71.5  \\
Hypergraph RAG w/ Visualization & 79.0 \\
\hline
\end{tabular}%
}
\vspace{-6pt}
\caption{Structural disambiguation results on lexically matched hard negative samples.}
\label{tab:adv_disambiguation}
\vspace{-12pt}
\end{table}

\noindent{\bf Enhancing Multi-hop Reasoning Composition.}  
To assess whether hypergraph visualization improves multi-hop reasoning, we evaluate 900 questions (evenly split across 2-hop, 3-hop, and 4-hop) from MusiQue \cite{trivedi2022musique} and HotpotQA \cite{yang2018hotpotqa}. Using Exact Match (EM) as the metric, we compare the Hyper-RAG pipeline \cite{feng2025hyperrag} with and without visual hypergraph input. Results in Tab.~\ref{tab:adv_multihop} show consistent improvements across all reasoning depths, with the most significant gains in 3-hop (+6.5\%) and 4-hop (+10.6\%) tasks, highlighting the utility of hypergraph visualization in complex reasoning.

\begin{table}[t]
\centering
\vspace{-5pt}
\resizebox{0.95\linewidth}{!}{
\begin{tabular}{lcc}
\hline
Reasoning EM & w/o visualization & w/ visualization \\
\hline
2-hop Reasoning & 72.3 & 74.0 \\
3-hop Reasoning & 60.5 & 67.0 \\
4-hop Reasoning & 48.7 & 59.3 \\
\hline
\end{tabular}
}
\vspace{-8pt}
\caption{Visualization effects on multi-hop reasoning.}
\label{tab:adv_multihop}
\vspace{-5pt}
\end{table}

\noindent{\bf MLLM Hypergraph Perception Capability.}
We also evaluate the hypergraph perception capabilities of representative frozen off-the-shelf MLLMs through two key tasks: \textit{overlap-core identification} and \textit{bridge-hyperedge detection}. Both tasks (illustrated in Fig.~\ref{fig:bias}(b)) are conducted on a fixed 1,000-sample subset from the UltraDomain dataset. As shown in Tab.~\ref{tab:fea_model}, all MLLMs demonstrate notable accuracy in identifying these structural cues from hypergraphs, which serves as a strong foundation for integrating hypergraph visualization into their reasoning processes within RAG pipelines.

\begin{table}[t]
\centering
\resizebox{\linewidth}{!}{%
\begin{tabular}{lcc}
\hline
Frozen MLLM & Overlap-core Identification & Bridge Hyperedge Detection\\
\hline
Qwen3VL-8B & 86.8\% & 83.2\% \\
Gemini-3 Flash & 90.5\% & 88.9\% \\
\hline
\end{tabular}%
}
\vspace{-10pt}
\caption{Accuracy (\%) of off-the-shelf MLLMs on key structural detection tasks.}
\label{tab:fea_model}
\vspace{-19pt}
\end{table}

\noindent{\bf Efficiency Analysis.} We analyze the retrieval
\begin{wrapfigure}{r}{0.55\linewidth}
\centering
\vspace{-15pt}
\includegraphics[width=\linewidth]{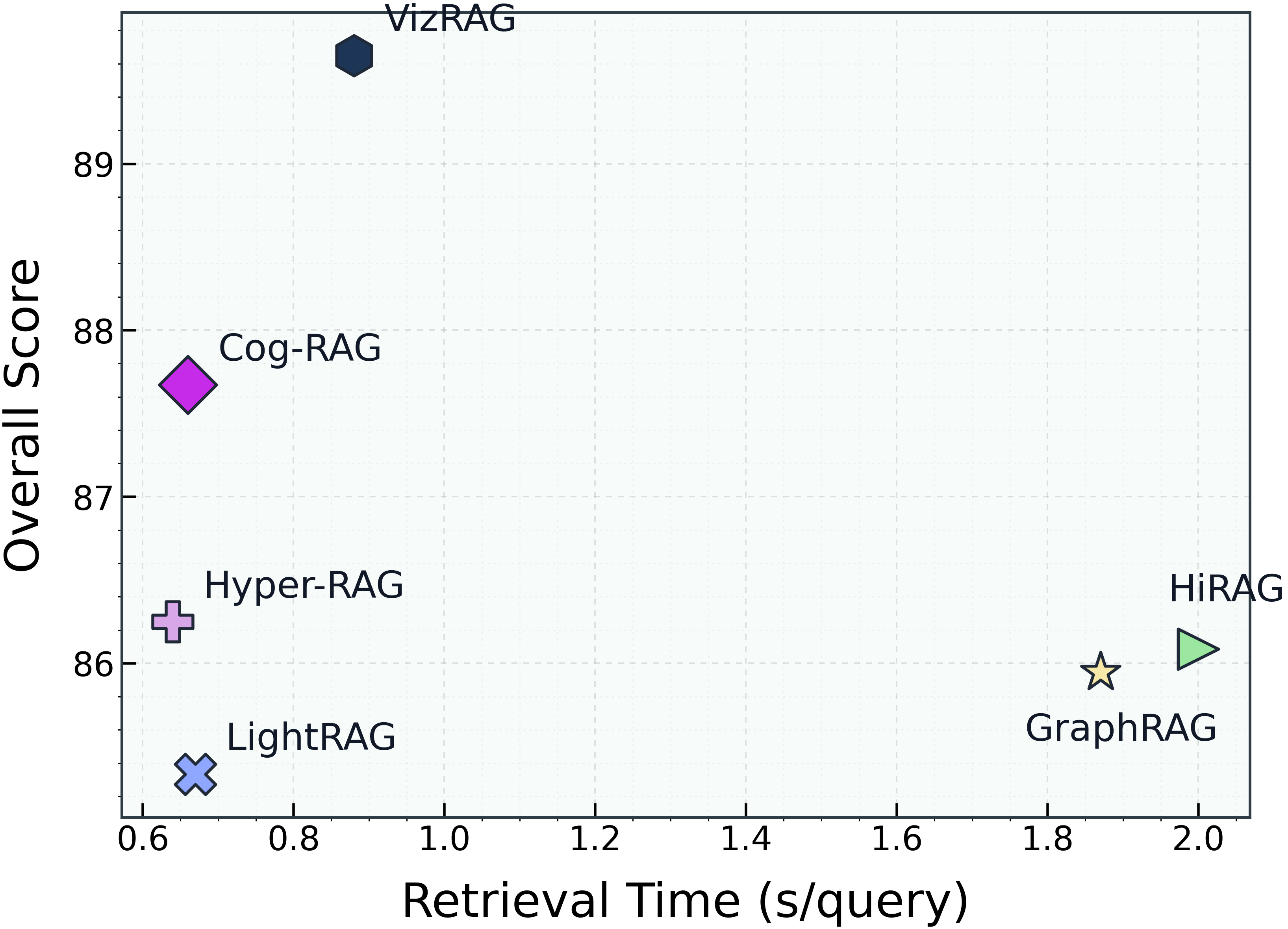}
\vspace{-22pt}
\caption{Trade-off between retrieval time and score.}
\label{fig:time_efficiency}
\vspace{-15pt}
\end{wrapfigure}
 efficiency for the VizRAG and baselines. As shown in Fig.~\ref{fig:time_efficiency}, VizRAG achieves the highest overall score with low retrieval time, striking a balance between efficiency and performance. The optimized visualization adds only 0.21 seconds per query, a lightweight overhead compared to the baseline, while yielding significant score gains (Tab.~\ref{tab2}). This makes VizRAG ideal for scenarios demanding both accuracy and speed.

\section{Conclusion}
We introduce hypergraph visualization as a novel enhancement for RAG systems, along with the plug-and-play HyperViz toolkit and VizRAG framework, enabling multimodal hypergraph reasoning and achieving significant gains in real-world knowledge-intensive tasks.

\section*{Limitations}
While VizRAG demonstrates promising results, it faces several limitations. Its performance heavily depends on the chosen MLLM's ability to perceive visual structures, which may limit its effectiveness in scenarios involving complex visual reasoning tasks. Rendering combined with dual-view probing introduces additional system overhead, particularly in latency-sensitive applications. Furthermore, the reliance on LLM-as-a-judge evaluations may lead to evaluator bias. Future work will focus on developing adaptive visualization strategies that dynamically adjust to task-specific requirements and incorporating broader, human-centered evaluation methods to reduce bias and improve robustness.


\bibliography{custom}

\clearpage
\appendix
\section{Related Works}
\label{sec:related}
\paragraph{RAG Systems.} Existing RAG systems can be categorized into three types based on their knowledge structures: (i) Document-based RAG methods \cite{asai2023self, zhang2024arl2, xia2025improving,yang2024crag} primarily rely on flat document structures for retrieval and generation. However, they struggle with modeling semantic relationships and contextual dependencies, limiting the accuracy and completeness of generated responses. (ii) GraphRAG methods \cite{sarmah2024hybridrag, peng2024graph} address the limitations by incorporating knowledge graphs, which enhance the organization and semantic expression of content retrieval by structurally representing entities and their relationships. Despite these advancements, GraphRAG often constructs sparse graph structures, failing to fully capture higher-order semantic associations and complex contextual relations. (iii) HypergraphRAG \cite{feng2025hyper, hu2026cog} extends this approach by utilizing hypergraphs, where hyperedges connect multiple nodes to represent higher-order semantic connections and group semantics, significantly enhancing the expressiveness of knowledge structures and retrieval quality. However, these HypergraphRAG methods solely rely on text input, which fail to adapt modern MLLMs' super vision capability, motivating our exploration of hypergraph visualization to enhance them.

\paragraph{Hypergraph Visualization.}
Existing hypergraph visualization methods can be broadly grouped into two categories. (i) \textbf{Set-based visualization methods} treat each hyperedge as a set and focus on overlap and intersection patterns. Representative designs include contour-based approaches such as Bubble Sets \cite{collins2009bubble}, matrix-based intersection analysis such as UpSet \cite{lex2014upset}, and element-attribute analysis for dense set systems such as GridSet \cite{chung2020gridset}. (ii) \textbf{Induced visualization methods} convert a hypergraph into an incidence (bipartite) graph by introducing virtual nodes for hyperedges \cite{bretto2013hypergraph}, and then apply mature graph visualization techniques \cite{graphviz,networkx} to the induced structure. The first preserves set semantics directly, while the second offers better compatibility with scalable graph layouts and interaction pipelines.

\section{HyperViz Toolkit Details}
\label{app:tool}

This appendix supplements the HyperViz toolkit's extra characteristics not covered in the main texts.

\subsection{Additional HyperViz Characteristics}
\label{app:chac}
\paragraph{Customizable Visual Styles.}
HyperViz offers extensive customization options during rendering, including adjustable parameters for node shapes, edge thickness, background colors, node outline styles, and graph margins. For example, node shapes can be set to options such as \texttt{ellipse}, \texttt{circle}, \texttt{box}, or \texttt{polygon}, with defaults like \texttt{ellipse} and \texttt{rounded box} commonly used to differentiate entities and hyperedges. Edge thickness can be customized within a range of 0.1 to 10 pixels, while background colors support RGB, HEX, or standard color names. Node outline styles include options such as \texttt{solid}, \texttt{dashed}, and \texttt{dotted}. The spacing between the graph and canvas edges can be adjusted within a range of 0 to 100 pixels. Additionally, in the set-based paradigm, hyperedge region colors can be selected from predefined palettes (e.g., \texttt{pastel} or \texttt{vivid}) or defined by the user. HyperViz also provides toggle options such as \texttt{show\_legend}, \texttt{show\_title}, \texttt{show\_footer}, and \texttt{show\_edge\_labels}, allowing users to control the display of legends, titles, footers, and edge labels. These features collectively enable HyperViz to adapt to a wide range of visualization tasks and requirements.

\paragraph{Interactive Visualization.}
Beyond static PNG outputs, HyperViz preserves stable visual IDs (entity IDs as \texttt{E1, E2,...}, hyperedge IDs as \texttt{H1, H2,...}) and exports synchronized graph-text mappings, which support interactive front-end behaviors without changing backend retrieval logic.
In practical deployment, we bind these IDs to metadata tables and enable interaction primitives, including zoom, pan, hover-tooltip, and click-to-expand neighborhood.
Clicking an entity can trigger one-hop or multi-hop expansion through $\mathcal{N}(v,\mathcal{G})$, while clicking a hyperedge opens its member set, relation description, source chunk ID, and evidence text.
This ID-consistent design guarantees that interactive exploration and MLLM prompting refer to the same structural objects.

\paragraph{Scalability Optimization.}
HyperViz uses a staged compression-render pipeline for dense subgraphs.
First, it applies budgeted subgraph construction with explicit caps (e.g., \texttt{max\_edges} and \texttt{max\_entities}) and preserves high-utility structures through query-aware expansion.
Second, for panel rendering, it uses mode-specific simplification: induced layouts keeps explicit E-H participation edges, and set-based layouts emphasizes group semantics by contour aggregation.
Third, for multi-view usage, \texttt{render\_hypergraph\_collage} arranges complementary views in a bounded panel grid to avoid single-canvas overplotting.
Together with the main-text pruning strategy (importance-based filtering under a congestion budget), these controls preserve readability while retaining key high-order evidence.

More detailed characteristics and features of HyperViz will be compiled into a comprehensive manual, which will be made available on a dedicated website following the acceptance of the paper, along with the code.

\subsection{Hyperparameter Optimization Details}
\label{app:hyper}
We adopt a unified grid search strategy to optimize the full set of hyperparameters in HyperViz, including $w_v$, $w_e$, $\gamma$, $\alpha$, and $\beta$. Due to the inherent mutual dependencies across these hyperparameters, all parameters are jointly optimized on real-world knowledge hypergraphs, rather than adopting separate stage-wise tuning. All spatial weight parameters $w_v, w_e$ and structural regularization parameters $\gamma, \alpha, \beta$ are searched simultaneously within their respective ranges of $[1, 10]$ and $[0, 1]$.

The optimization is performed entirely on real knowledge hypergraphs constructed from the UltraDomain Legal dataset \cite{qian2025memorag}. To strictly prevent data leakage, this tuning subset is excluded from all evaluation test sets throughout our experiments. The adopted real legal hypergraphs naturally cover diverse density distributions, with node degrees ranging from $1$ to $9$ and hyperedge sizes spanning $[2, 11]$, fully encompassing sparse, moderate, and dense structural patterns in real document scenarios.

All grid search trials are supervised by the structural perception capability of Gemini-3 Flash, utilizing the overlap-core identification and bridge hyperedge detection tasks illustrated in Fig.~\ref{fig:bias}. The final parameter combination is selected to maximize structural perception accuracy, which we equate to visual readability for MLLM-centric visualization evaluation.
\subsection{Description of Layouts in Hyperviz}
\label{app:intro_layout}

HyperViz provides six layout variants under two complementary paradigms: set-based views for group semantics and Induced views for explicit incidence relations. In implementation, these layouts are achieved by customizing layout algorithms in Graphviz \cite{graphviz} with paradigm-specific post-processing (e.g., contour generation for set regions and bipartite incidence expansion for induced views).

Set-based layouts represent high-order relationships (i.e., hyperedges) as set elements through region containment to highlight entity grouping relations. In Hyperviz, it is implemented in three diverse layout variants:
\begin{itemize}
    \item \noindent\textbf{Set-Hier layout} is achieved by the Graphviz \texttt{dot} layout algorithm. It emphasizes top-down hierarchy and dependency chains, making parent-child style organization and progressive evidence flow easier to inspect.
    \item \textbf{Set-Radial layout} is achieved by the Graphviz \texttt{twopi} layout algorithm. It emphasizes center-periphery organization around high-degree entities, which is useful for identifying hubs, overlap cores, and radius-like influence patterns.
    \item \noindent\textbf{Set-Force layout} is achieved by the Graphviz \texttt{fdp}/\texttt{sfdp} layout algorithms. It emphasizes community structure and local cluster separation, helping reveal dense semantic groups and weak inter-group bridges.
\end{itemize}

Induced layouts emphasize direct element-group relationships by representing hyperedges as special nodes, with pairwise edges explicitly connecting these nodes to their corresponding hyperedge nodes. 

\begin{itemize}
    \item \noindent\textbf{Bi-Inc:} uses NetworkX \texttt{bipartite\_layout} to separate entity nodes and virtual hyperedge nodes into two partitions, making membership links directly readable. Incidence edges are drawn with light dashed strokes, entity nodes are styled as blue-outlined circles, and virtual hyperedge nodes are styled as orange-outlined squares.
    \item \noindent\textbf{Ra-Cen:} first selects the center as the highest-degree entity in the induced graph, then computes shortest-path distance shells and renders them using NetworkX \cite{networkx} \texttt{shell\_layout}. This design emphasizes center-periphery organization for neighborhood expansion and multi-hop anchor tracing; if no valid center is available, it falls back to \texttt{spring\_layout}.
    \item \noindent\textbf{Cli-Exp:} keeps the induced incidence backbone and additionally expands each hyperedge $e = \{v_1, v_2, \ldots, v_k\}$ into pairwise entity links (i.e., $\binom{k}{2}$ clique edges). The panel is laid out by NetworkX \texttt{spring\_layout} (\texttt{seed}=42, $k=0.95$), where dashed incidence edges preserve E--H semantics and solid clique edges improve local connectivity perception.
\end{itemize}

To maximize the interpretability of visualized hypergraphs, explicit semantic annotations are embedded:
{\bf (1) Node-Level Annotation.} Intuitive text labels are attached to all entity nodes, presenting clear semantic information. {\bf (2) Hyperedge-Level Annotation.} Internal inclusion relations of hyperedges are summarized and displayed via unified legend labels, clarifying hierarchical associations.

\section{Dataset Statistics}
\label{app:data}

\begin{table}[h]
    \centering
    \resizebox{\linewidth}{!}{
    \begin{tabular}{lccccc}
    \toprule
    \textbf{Statistics} & \textbf{Mix} & \textbf{CS} & \textbf{Agriculture}  & \textbf{Neurology} & \textbf{Pathology} \\
    \midrule
    Total Documents & 61 & 10 & 12 & 1 & 1 \\
    Total Chunks &  560 & 1992 & 1813 & 1790 & 824 \\
    Total Tokens & 615,355 & 2,190,803 & 1,993,515 & 1,968,716 & 905,760 \\
    \bottomrule
    \end{tabular}}
    \caption{Statistical information of the reported datasets.}
    \label{tab:app_data_stats}
\end{table}

Table~\ref{tab:app_data_stats} reports the dataset statistics used in this paper.

\section{Baseline Details}
\label{app:baseline}

We compare VizRAG with representative methods spanning graph and hypergraph RAG families.

\noindent\textbf{GraphRAG} \cite{edge2024local}: entity-relation graph construction with graph-guided retrieval and community-level summaries.

\noindent\textbf{LightRAG} \cite{guo2024lightrag}: lightweight graph-enhanced retrieval with dual-level graph context integration.

\noindent\textbf{HiRAG} \cite{huang2025retrieval}: hierarchical graph retrieval with multi-level structural traversal.

\noindent\textbf{Hyper-RAG} \cite{feng2025hyper}: hypergraph retrieval modeling high-order relations through hyperedges.

\noindent\textbf{Cog-RAG} \cite{hu2026cog}: cognitive-inspired dual-hypergraph retrieval with theme-aligned top-down evidence activation.

For fairness, we keep the chunking strategy, embedding model, and evaluation protocol aligned with Cog-RAG \cite{hu2026cog} and keep consistent across methods.
\begin{figure*}[t]
    \centering
    \includegraphics[width=\linewidth]{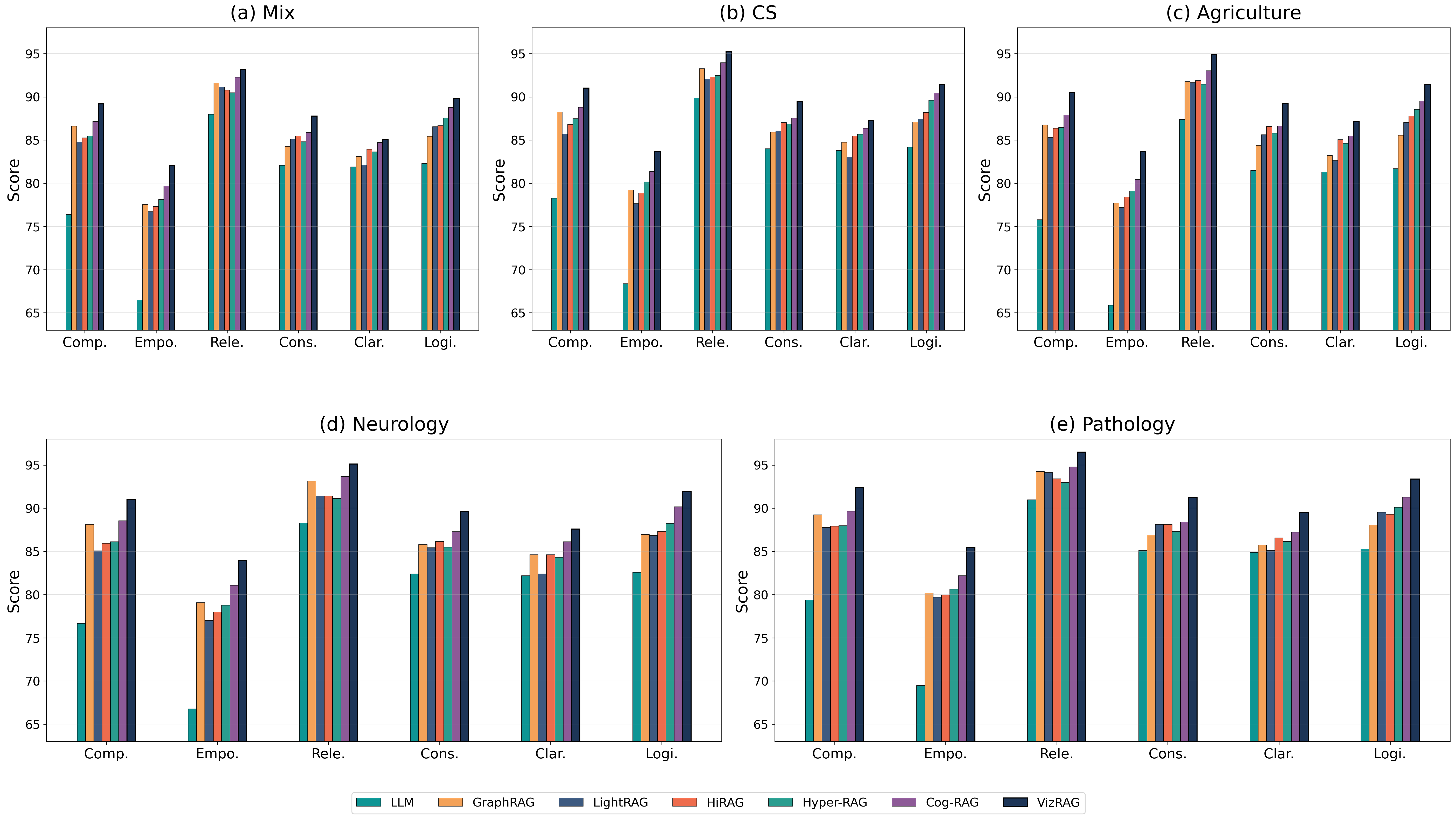}
    \caption{Score-based metric comparison of full five datasets.
}
    \label{fig:indi_datascore}
    \vspace{-15pt}
\end{figure*}
\section{Evaluation Protocol Details}
\label{app:eval}
We evaluated the proposed method and baseline across six key dimensions: Comprehensiveness, Empowerment, Relevance, Consistency, Clarity, and Logical Coherence.

\subsection{Selection-based Evaluation}
For qualitative comparison, we report win rates between methods. To ensure fairness, the order of answers in each method pair was alternated in the prompts, and the average result was calculated. The evaluation dimensions are described below:

\begin{itemize}
    \item \textbf{Comprehensiveness (Comp.)}: Measures how thoroughly the answer addresses all aspects and details of the question.
    \item \textbf{Empowerment (Empo.)}: Assesses the extent to which the answer helps the reader understand the topic and make informed decisions.
    \item \textbf{Relevance (Rele.)}: Evaluates how accurately the response focuses on the key aspects of the question without including irrelevant information.
    \item \textbf{Consistency (Cons.)}: Examines the ability of the system to synthesize and integrate information from multiple sources into a coherent and logical response.
    \item \textbf{Clarity (Clar.)}: Gauges whether the response avoids unnecessary verbosity and redundancy while providing complete and easily understandable information.
    \item \textbf{Logical Coherence (Logi.)}: Checks whether the response maintains consistent and rational arguments without contradictions throughout.
\end{itemize}

\subsection{Score-based Evaluation}
For quantitative assessment, we employed LLMs to score the responses based on the following indicators:

\begin{itemize}
    \item \textbf{Comprehensiveness (0-100)}: Evaluates whether the response comprehensively covers all critical aspects of the question and identifies any omissions.
    \item \textbf{Empowerment (0-100)}: Measures the credibility and persuasiveness of the response. High scores are given to answers that provide strong evidence or cite authoritative sources.
    \item \textbf{Relevance (0-100)}: Determines the degree to which the response stays focused on the question and avoids unnecessary digressions.
    \item \textbf{Consistency (0-100)}: Assesses whether the response is logically organized, flows smoothly, and the components are well-connected and mutually supportive.
    \item \textbf{Clarity (0-100)}: Examines whether the response is expressed clearly and concisely, using precise language and unambiguous definitions.
    \item \textbf{Logical Coherence (0-100)}: Evaluates whether the response maintains logical consistency and is free from contradictions.
\end{itemize}

\subsection{Rating Levels}
Each evaluation dimension is assessed using five discrete levels, with clear scoring criteria to ensure consistency and transparency. For example, the \textbf{Comprehensiveness} dimension is rated as follows:

\begin{itemize}
    \item \textbf{Level 1 (0-20)}: The response is highly incomplete and omits significant portions or key aspects of the question.
    \item \textbf{Level 2 (20-40)}: The response includes some relevant content but misses many important elements, making it insufficiently comprehensive.
    \item \textbf{Level 3 (40-60)}: The response addresses the main aspects of the question but still has notable omissions.
    \item \textbf{Level 4 (60-80)}: The response is mostly comprehensive, covering most aspects of the question with minimal omissions.
    \item \textbf{Level 5 (80-100)}: The response is exceptionally thorough, addressing all aspects of the question without any omissions, providing a complete understanding to the reader.
\end{itemize}

\section{Additional Results}
\label{app:more_results}
\subsection{Detailed Scores on Individual Datasets}
Fig.\ref{fig:indi_datascore} shows the results of scores under different evaluation dimensions on the individual Mix, CS, Agriculture, Neurology, and Pathology datasets.

\section{Prompt Templates}
\label{app:prompt}
\subsection{Core Prompt Set Aligned with VizRAG}

\begin{myprompt_single}{Extracting Entities}
    \textbf{Formulation}: $\mathcal{P}_\text{ext\_entity}(D_i)$\newline\newline
    \textbf{Prompt: }\textit{
    You are given one chunk $D_i$.
    Extract all important entities for retrieval and reasoning.
    Return JSON only with the schema:
    \newline
    \{ "entities": [
      \{ "entity\_name": str,
         "entity\_type": one of [organization, person, geo, event, role, concept, method, artifact, disease, treatment, mechanism],
         "entity\_description": str,
         "aliases": [str],
         "source\_spans": [str]
      \}
    ] \}
    \newline
    Requirements:
    \newline
    1) Use the same language as the input chunk.
    \newline
    2) Merge duplicates and aliases.
    \newline
    3) Keep entities that are structurally useful for downstream hyperedge construction.
}
\end{myprompt_single}

\begin{myprompt_single}{Extracting Hyperedges}
    \textbf{Formulation}: $\mathcal{P}_\text{ext\_relation}(D_i,\mathcal{V}_{D_i})$\newline\newline
    \textbf{Prompt: }\textit{
    You are given a chunk $D_i$ and the extracted entity set $\mathcal{V}_{D_i}$.
    Construct hyperedges that represent semantically coherent multi-entity relations (including pairwise entity relationships as a special case).
    Return JSON only with the schema:
    \newline
    \{ "hyperedges": [
      \{ "entities\_set": [entity\_name],
         "relation\_description": str,
         "relation\_keywords": [str],
         "relation\_strength": number in [0,100],
         "evidence\_spans": [str]
      \}
    ] \}
    \newline
    Requirements:
    \newline
    1) Each hyperedge must contain at least 2 (including 2) valid entities from $\mathcal{V}_{D_i}$.
    \newline
    2) Prefer compact but meaningful sets; avoid trivial or redundant hyperedges.
    \newline
    3) If two hyperedges are near-duplicates, keep the one with clearer evidence.
}
\end{myprompt_single}

\begin{myprompt_single}{Extracting Theme with Visual Cue}
    \textbf{Formulation}: $\mathcal{P}_\text{ext\_theme}(D_i,\mathcal{I}_{D_i})$\newline\newline
    \textbf{Prompt: }\textit{
    You are given chunk text $D_i$ and its hypergraph image $\mathcal{I}_{D_i}$.
    This visual input $\mathcal{I}_{D_i}$ encodes the intra-chunk knowledge structure over entity objects and their grouping membership.
    Infer a concise theme that is consistent with both text and structure.
    Return JSON only:
    \newline
    \{ "theme\_description": str,
       "support\_entity\_ids": ["E*"],
       "support\_hyperedge\_ids": ["H*"]
    \}
    \newline
    Requirements:
    \newline
    1) Theme must reflect the central semantic focus of the chunk.
    \newline
    2) Prioritize evidence supported by both text and visual structure.
}
\end{myprompt_single}

\begin{myprompt_single}{Extracting Theme-related Key Entities with Visual Cue}
    \textbf{Formulation}: $\mathcal{P}_\text{ext\_key}(\mathcal{E}^{theme}_{D_i},D_i,\mathcal{I}_{D_i},\mathcal{V}_{D_i})$\newline\newline
    \textbf{Prompt: }\textit{
    You are given: (a) theme description, (b) chunk text $D_i$, (c) hypergraph image $\mathcal{I}_{D_i}$ with stable IDs, and (d) candidate entities $\mathcal{V}_{D_i}$.
    This visual input $\mathcal{I}_{D_i}$ encodes the intra-chunk knowledge structure over entity objects and their grouping membership.
    Select theme-critical key entities for top-layer indexing.
    Return JSON only:
    \newline
    \{ "key\_entities": [
      \{ "entity\_name": str,
         "entity\_id": "E*" (if available),
         "key\_entity\_type": str,
         "key\_entity\_description": str,
         "key\_score": number in [0,100],
         "selection\_reason": str
      \}
    ] \}
    \newline
    Requirements:
    \newline
    1) Prioritize entities that are both theme-relevant and structurally central.
    \newline
    2) Avoid selecting many near-duplicate entities.
}
\end{myprompt_single}

\begin{myprompt_single}{Extracting Query Keywords for Top Retrieval}
    \textbf{Formulation}: $\mathcal{P}_\text{keyword}(q)$\newline\newline
    \textbf{Prompt: }\textit{
    Given user query $q$, extract retrieval keywords for top-layer hyperedge search.
    Return JSON only:
    \newline
    \{ "theme\_keywords": [str], "entity\_keywords": [str] \}
    \newline
    Requirements:
    \newline
    1) Keep keywords short and retrieval-oriented.
    \newline
    2) Cover both abstract intent (theme-level) and concrete anchors (entity-level).
    \newline
    3) Remove redundant synonyms when possible.
}
\end{myprompt_single}

\begin{myprompt_single}{Locating Query-relevant Entities from Top-view Visualization}
    \textbf{Formulation}: $\mathcal{P}_\text{locate}(q,\mathcal{I}_{Top},\mathcal{C}_{e\_rel})$\newline\newline
    \textbf{Prompt: }\textit{
    You are given query $q$, top-layer subgraph visualization $\mathcal{I}_{Top}$, and context summaries $\mathcal{C}_{e\_rel}$ of retrieved theme hyperedges.
    The visual input $\mathcal{I}_{Top}$ encodes top-layer knowledge structure, including entities, theme hyperedges, and their structural connectivity.
    Identify entities in the visual structure that are most relevant for downstream bottom retrieval.
    Return JSON only:
    \newline
    \{ "selected\_entities": [
      \{ "entity\_id": "E*",
         "entity\_name": str,
         "relevance\_score": number in [0,100],
         "reason": str
      \}
    ] \}
    \newline
    Requirements:
    \newline
    1) Use both visual structure (bridges, overlaps, hubs) and textual context.
    \newline
    2) Prefer entities that can connect to answer-critical evidence paths.
}
\end{myprompt_single}

\begin{myprompt_single}{Final Answer Generation with Bottom-view Evidence}
    \textbf{Formulation}: $\mathcal{P}_\text{answer}(q,\mathcal{I}_{Bot},\mathcal{V}_{rel},\mathcal{E}_{dif},\mathcal{C}_{v\_rel},\mathcal{C}_{e\_dif})$\newline\newline
    \textbf{Prompt: }\textit{
    You are given query $q$, bottom-layer visualization $\mathcal{I}_{Bot}$, retrieved entities $\mathcal{V}_{rel}$, expanded hyperedges $\mathcal{E}_{dif}$, and their textual contexts $\mathcal{C}_{v\_rel}, \mathcal{C}_{e\_dif}$.
    The visual input $\mathcal{I}_{Bot}$ encodes bottom-layer evidence structure over entities, hyperedges, and their local reasoning paths.
    Generate the final answer grounded in the provided evidence.
    Return JSON only:
    \newline
    \{ "answer": str,
       "support\_entity\_ids": ["E*"],
       "support\_hyperedge\_ids": ["H*"],
       "confidence": number in [0,100]
    \}
    \newline
    Requirements:
    \newline
    1) Prefer evidence-consistent reasoning over unsupported speculation.
    \newline
    2) If evidence is insufficient, explicitly state uncertainty.
    \newline
    3) Keep the answer concise but complete for the question.
}
\end{myprompt_single}

\subsection{Evaluation Metrics}
\begin{myprompt_double}{Selection-based Evaluation}
    \textbf{Formulation}: $\mathcal{P}_\text{eval\_scoring}(q, \mathcal{A}_1, \mathcal{A}_2)$\newline 
    $q$ denotes user query, $\mathcal{A}_1$ and $\mathcal{A}_2$ denotes the response from two approaches. \newline\newline
    \textbf{Prompt: }\textit{You will evaluate two answers to the same question based on six criteria: Comprehensiveness, Empowerment, Relevance, Consistency, Clarity, and Logical.\newline
    ---Goal---\newline
    You will evaluate two answers to the same question by using the relevant documents based on six criteria: Comprehensiveness, Empowerment, Relevance, Consistency, Clarity, and Logical.\newline
    -Comprehensiveness: How much detail does the answer provide to cover all aspects and details of the question?\newline
    -Empowerment: How well does the answer help the reader understand and make informed judgments about the topic?\newline
    ...,\newline
    -Logical: How well does the system maintain consistent logical arguments without contradicting itself across the response?\newline\newline
    For each criterion, choose the better answer (either Answer 1 or Answer 2) and explain why. Then, select an overall winner based on these six categories.\newline
    Here is the question: $q$\newline
    Here are the two answers: \newline
    Answer 1: $\mathcal{A}_1$;\newline
    Answer 2: $\mathcal{A}_2$\newline
    Evaluate both answers using the six criteria listed above and provide detailed explanations for each criterion.
}
\end{myprompt_double}

\clearpage

\begin{myprompt_double}{Scoring-based Evaluation}
    \textbf{Formulation}: $\mathcal{P}_\text{eval\_scoring}(q, \mathcal{A}, \mathcal{C})$\newline 
    $q$ denotes user query, $\mathcal{A}$ denotes LLM response, $\mathcal{C}$ denotes the original text chunk that generated the question. \newline\newline
    \textbf{Prompt: }\textit{You are an expert tasked with evaluating answers to the questions by using the relevant documents based on five criteria: Comprehensiveness, Diversity, Empowerment, Logical, and Readability.\newline\newline
    ---Goal---\newline
     You will evaluate the answers to the question by using the relevant documents based on six criteria: Comprehensiveness, Empowerment, Relevance, Consistency, Clarity, and Logical.\newline\newline
    -Comprehensiveness-\newline
    Measure whether the answer comprehensively covers all key aspects of the question and whether there are omissions.\newline
    Level   $|$ score range $|$ description\newline
    Level 1 $|$ 0-20   $|$ The answer is extremely one-sided, leaving out key parts or important aspects of the question.\newline
    Level 2 $|$ 20-40  $|$ The answer has some content, but it misses many important aspects of the question and is not comprehensive enough.\newline
    Level 3 $|$ 40-60  $|$ The answer is more comprehensive, covering the main aspects of the question, but there are still some omissions.\newline
    Level 4 $|$ 60-80  $|$ The answer is comprehensive, covering most aspects of the question, with few omissions.\newline
    Level 5 $|$ 80-100 $|$ The answer is extremely comprehensive, covering all aspects of the question with no omissions, enabling the reader to gain a complete understanding.\newline
    ...,\newline\newline
    For each indicator, please give the problem a corresponding Level based on the description of the indicator, and then give a score according to the score range of the level.\newline
    Here is the question: $q$\newline
    Here are the relevant document: $\mathcal{C}$\newline
    Here are the answer: $\mathcal{A}$\newline
    Evaluate all the answers using the six criteria listed above. For each criterion, provide a summary description, give a Level based on the description of the indicator, and then give a score based on the score range of the level.
}
\end{myprompt_double}

\end{document}